# Lensless Compressive Sensing Imaging

Gang Huang, Hong Jiang, Kim Matthews and Paul Wilford

*Abstract*—In this paper, we propose a lensless compressive sensing imaging architecture. The architecture consists of two components, an aperture assembly and a sensor. No lens is used. The aperture assembly consists of a two dimensional array of aperture elements. The transmittance of each aperture element is independently controllable. The sensor is a single detection element, such as a single photo-conductive cell. Each aperture element together with the sensor defines a cone of a bundle of rays, and the cones of the aperture assembly define the pixels of an image. Each pixel value of an image is the integration of the bundle of rays in a cone. The sensor is used for taking compressive measurements. Each measurement is the integration of rays in the cones modulated by the transmittance of the aperture elements. A compressive sensing matrix is implemented by adjusting the transmittance of the individual aperture elements according to the values of the sensing matrix. The proposed architecture is simple and reliable because no lens is used. Furthermore, the sharpness of an image from our device is only limited by the resolution of the aperture assembly, but not affected by blurring due to defocus. The architecture can be used for capturing images of visible lights, and other spectra such as infrared, or millimeter waves. Such devices may be used in surveillance applications for detecting anomalies or extracting features such as speed of moving objects. Multiple sensors may be used with a single aperture assembly to capture multi-view images simultaneously. A prototype was built by using a LCD panel and a photoelectric sensor for capturing images of visible spectrum.

*Index Terms*— Compressive sensing, imaging, lensless, sensor

## I. Introduction

COMPERSSIVE sensing [1][2] is an emerging technique to acquire and process digital data such as images and videos [3][4][5][6]. Compressive sensing is most effective when it is used in data acquisition: to capture the data in the form of compressive measurements [7]. With compressive measurements, images may be reconstructed with far fewer measurements than the number of pixels in the original images. Therefore, by using compressive sensing in acquisition, images are compressed while they are captured, avoiding high speed processing, or transmission, of a large number of pixels.

The first device that directly captures compressive measurements of an image is the single pixel camera of [8][9]. It is a camera architecture that employs a digital micromirror array to perform optical calculations of linear projections of an image onto pseudorandom binary patterns. It has the ability to obtain an image with a single detection element while sampling the image fewer times than the number of pixels. The same camera architecture is also used for Terahertz imaging [10][11], and millimeter wave imaging [12]. These cameras all make use of a lens to form an image in a plane before the image is projected onto a pseudorandom binary pattern. Lenses, however, severely constrain the geometric and radiometric mapping from the scene to the image [13]. Furthermore, lenses add size, cost and complexity to a camera.

In this paper, we propose architecture for compressive sensing imaging without a lens. The proposed architecture consists of two components, an aperture assembly and a single sensor. No lens is used. The aperture assembly consists of a two dimensional array of aperture elements. The transmittance of each aperture element is independently controllable. The sensor is a single detection element, such as a single photo-conductive cell. Each aperture element together with the sensor defines a cone of a bundle of rays, and the cones of the aperture assembly define the pixels of an image. The sensor is used for taking compressive measurements. Each measurement is the integration of rays in the cones modulated by the transmittance of the aperture elements.

The proposed architecture is different from the cameras of [8] and [13]. The fundamental difference is how the image is formed. In both [8] and [13], an image of the scene is formed on a plane, by some physical mechanism such a lens or a pinhole, before it is digitally captured (by compressive measurements in [8], and by pixels in [13]). In the proposed architecture of this work, no image is physically formed before the image is captured. The detailed discussion on the difference will be given in Section III.

The proposed architecture is distinctive with the following features.

- No lenses are used. An imaging device using the proposed architecture can be built with reduced size, weight, cost and complexity. In fact, our architecture does not rely on any physical mechanism to form an image before it is digitally captured.
- No scene is out of focus. The sharpness and resolution of images from the proposed architecture are only limited by the resolution of the aperture assembly (number of aperture elements), there is no blurring introduced by lens for scenes that are out of focus.
- Multi-view images can be captured simultaneously by a device using multiple sensors with one aperture assembly.
- The same architecture can be used for imaging of visible spectrum, and other spectra such as infrared and millimeter waves.

The authors are with Bell Labs, Alcatel-Lucent, 700 Mountain Ave, Murray Hill, NJ 07974. Emails: firstname.lastname@alcatel-lucent.com

- Devices based on this architecture may be used in surveillance applications [6] for detecting anomalies or extracting features such as speed of moving objects.

We built a prototype device for capturing images of visible spectrum. It consists of an LCD panel, and a sensor made of a three-color photo-electric detector.

The organization of this paper is as follows. In the next section, the architecture of our work is described. The related work is discussed in Section III. The mathematical formulation for images of the proposed architecture is given in Section IV, followed by a discussion, in Section V, of multi-view imaging by using multiple sensors with one aperture assembly. In Section VI, issues arising from practical implementations of the architecture are addressed. The prototype system is described in Section VII.

## II. DESCRIPTION OF ARCHITECTURE

The proposed architecture is shown in Figure 1. It consists of two components: an aperture assembly and a sensor. The aperture assembly is made up of a two dimensional array of aperture elements. The transmittance of each aperture element, $T_{ij}$, can be individually controlled. The sensor is a single detection element, which is ideally of an infinitesimal size.

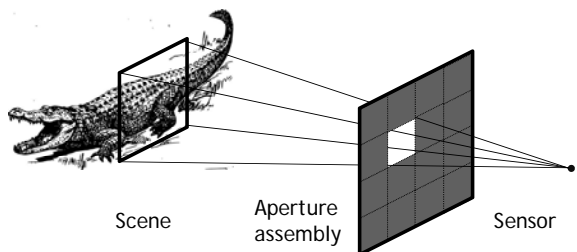

**Figure 1. The proposed architecture. It consists of two components: an aperture assembly and an infinitesimal sensor of a single detection element. Each element in the aperture assembly together with the sensor forms a cone of a bundle of rays, and the cones form the pixels of an image**

Each element of the aperture assembly, together with the sensor, defines a cone of a bundle of rays, see Figure 1, and the cones from all aperture elements are defined as pixels of an image. The integration of the rays within a cone is defined as a pixel value of the image. Therefore, in the proposed architecture, an image is defined by the pixels which correspond to the array of aperture elements in the aperture assembly.

An image can be captured by using the sensor to take as many measurements as the number of pixels. For example, each measurement can be made from reading of the sensor when one of the aperture elements is completely open and all others are completely closed, which corresponds to the binary transmittance $T_{ij} = 1$ (open), or $0$ (closed). The measurements are the pixel values of the image when the elements of the aperture assembly are opened one by one in certain scan order. This way of making measurements corresponds to the traditional representation of a digital image pixel by pixel. In the following, we describe how compressive measurements can be made in the proposed architecture.

### A. Compressive measurements

With compressive sensing, it is possible to represent an image by using fewer measurements than the number of pixels [3][4][5][6]. The architecture of Figure 1 makes it simple to take compressive measurements.

To make compressive measurements, a sensing matrix is first defined. Each row of the sensing matrix defines a pattern for the elements of the aperture assembly, and the number of columns in a sensing matrix is equal to the number of total elements in the aperture assembly. In the context of compressive sensing, the two dimensional array of aperture elements in the aperture assembly is conceptually rearranged into a one dimensional array, which can be done, for example, by ordering the elements of the aperture assembly one by one in certain scan order. Each value in a row of the sensing matrix is used to define the transmittance of an element of the aperture assembly. A row of the sensing matrix therefore completely defines a pattern for the aperture assembly, and it allows the sensor to make one measurement for the given pattern of the aperture assembly. The number of rows of the sensing matrix is the number of measurements, which is usually much smaller than the number of aperture elements in the aperture assembly (the number of pixels).

Let the sensing matrix be a random matrix whose entries are random numbers between 0 and 1. To make a measurement, the transmittance, $T_{ij}$, of each aperture element is controlled to equal the value of the corresponding entry in a row of the sensing matrix. The sensor integrates all rays transmitted through the aperture assembly. The intensity of the rays is modulated by the transmittances before they are integrated. Therefore, each measurement from the sensor is the integration of the intensity of rays through the aperture assembly multiplied by the transmittance of respective aperture element. A measurement from the sensor is hence a projection of the image onto the row of the sensing matrix. This is illustrated in Figure 2.

By changing the pattern of the transmittance of the aperture assembly, it is possible to make compressive measurements corresponding to a given sensing matrix whose entries have real values between 0 and 1.

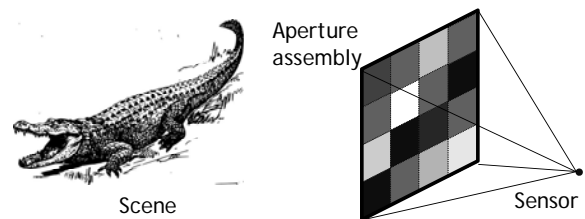

**Figure 2. Programmed aperture assembly for compressive measurements. The transmittances of aperture elements are controlled to match the values of a row of the sensing matrix. A measurement is the integration of all rays through the aperture assembly modulated by the transmittance values.**

## III. RELATED WORK

The proposed architecture is related to the single pixel camera of [8], which captures compressive measurements but has lenses, and the lensless camera of [13], which has no lenses but captures image pixels. At the first glance, our proposed architecture is simply a hybrid of the two; indeed, as far as the components and functionality are concerned, our architecture seems as if taking the lenses out of the camera of [8], or adding the projecting functionality into the camera of [13]. However, there is a fundamental difference between the architecture of this paper and the cameras of [8] and [13], which is how the images are formed. In both [8] and [13], a physical mechanism is used to form an image of the scene on a plane, and then the image on the plane is pixelized. In [8], a lens is employed to form an image of the scene on the micromirror array. The micromirror array then performs the functions of both pixelization and projection. In [13], attenuating aperture layers are used to create a pinhole which forms an image of the scene on the sensor array. The sensor array then pixelizes the pinhole image. Therefore, both cameras of [8] and [13] create an "analog" image of the scene on a plane.

In the cameras of [8] and [13], there are two processes that may affect the quality, sharpness and resolution, of an image. The first is the formation of the "analog" image on the plane of pixelization, and the second is the pixelization of the "analog" image. The former depends on the mechanism for forming the image. For example, in camera of [8], the sharpness may depend on the focal point of the scene, so that an object may appear blurred because it is out of focus. Furthermore, the artifact of blurring can occur even with theoretically perfect lens, micromirrors and sensor.

In the architecture of this work, no planar image is explicitly formed. One could argue that each measurement from the sensor is a projection of an image on the aperture assembly. However, this virtual image is not formed by any physical mechanism, and therefore, it is an ideal image that is free of any artifact such as blurring due to defocus. Therefore, the quality of image from the architecture of this work is only affected by the resolution of pixelization (the number of the aperture elements in the aperture assembly) if the aperture assembly and the sensor is theoretically perfect.

## IV. MATHEMATICAL FORMULATION

In this section, we formally define what an image is in the proposed architecture and how it is related to the measurements from the sensor. In particular, we will describe how a pixelized image can be reconstructed from the measurements taken from the sensor.

### A. Virtual image

Let the aperture assembly be a rectangular region on a plane with $(x, y)$ coordinate system. For each point, $(x, y)$, on the aperture assembly, there is a ray starting from a point on the scene, passing through the point $(x, y)$, and ending at the sensor, as shown in Figure 3. Therefore, there is a unique ray associated with each point $(x, y)$ on the aperture assembly, and its intensity arriving at the aperture assembly at time $t$ is denoted by $r(x, y; t)$. Then an image $I(x, y)$ of the scene is defined as the integration of the ray in a time interval $\Delta t$:

$$I(x, y) = \int_0^{\Delta t} r(x, y; t) dt. \quad (1)$$

Note that although the definition of an image in (1) is defined on the region of the aperture assembly, there is not an actual image physically formed in the architecture of this work. For this reason, the image of (1) is called a virtual image. A virtual image $I(x, y)$ can be considered as an analog image because it is continuously defined in the region of the aperture assembly.

Let the transmittance of the aperture assembly be define as $T(x, y)$. A measurement made by the sensor is the integration of the rays through the aperture assembly modulated by the transmittance, and it is given by

$$z_T = \iint T(x, y) I(x, y) dxdy. \quad (2)$$

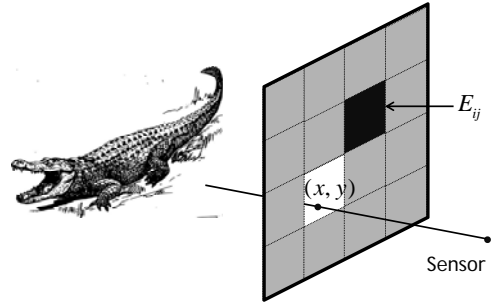

**Figure 3. A ray is defined for each point on the region of aperture assembly.**

Although the virtual image discussed above is defined on the plane of the aperture assembly, it is not necessary to do so. The virtual image may be defined on any plane that is placed in between the sensor and the aperture assembly and parallel to the aperture assembly.

### B. Pixelized image

The virtual image defined by (1) can be pixelized by the aperture assembly. Let the region defined by one aperture element be denoted by $E_{ij}$ as shown in Figure 3. Then the pixel value of the image at the pixel $(i, j)$ is the integration of the rays passing through the aperture element $E_{ij}$ and it is given by

$$I(i, j) = \iint_{E_{ij}} I(x, y) dxdy, \\ \iint \mathbf{1}_{E_{ij}}(x, y) I(x, y) dxdy. \quad (3)$$

In above, the function $\mathbf{1}_{E_{ij}}$ is the characteristic function of the aperture element $E_{ij}$. The characteristic function of a region $R$ is defined as

$$\mathbf{1}_R(x, y) = \begin{cases} 1, & (x, y) \in R \\ 0 & (x, y) \notin R \end{cases}. \quad (4)$$

Note that we use $I(i, j)$ to denote a pixelized image of a virtual image $I(x, y)$ which is analog.

Equation (3) defines the pixelized image $I(i, j)$. In compressive sensing, it is often mathematically convenient to reorder a pixelized image which is a two dimensional array into a one dimensional vector. Let $q$ be a mapping from a 2D array to a 1D vector defined by

$$q: (i, j) \mapsto n, \text{ so that } I_n = I(i, j). \quad (5)$$

Then the pixelized image $I(i, j)$ can be represented as a vector whose components are $I_n$. We will simply use $I$ to denote the pixelized image, either as a two dimensional array, or a one dimensional vector, interchangeably.

### C. Compressive measurements and reconstruction

When the aperture assembly is programmed to implement a compressive sensing matrix, the transmittance $T_{ij}$ of each aperture element is controlled to equal the value of the corresponding entry in the sensing matrix. For the $m$th measurement, the entries in row $m$ of the sensing matrix are used to program the transmittance of the aperture elements. Specifically, let the sensing matrix $A$ be a random matrix whose entries, $a_{mn}$, are random numbers between 0 and 1. Let $T_{ij}^m(x, y)$ be the transmittance of aperture element $E_{ij}$ for the $m$th measurement. Then, for the $m$th measurement, the transmittance of the aperture assembly is given by

$$T^m(x, y) = \sum_{i,j} T_{ij}^m(x, y), \text{ where} \quad (6)$$
$$T_{ij}^m(x, y) = a_{m,q(i,j)} \mathbf{1}_{E_{ij}}(x, y).$$

Therefore, according to (2), the measurements are given by

$$\begin{aligned} z_m &= \iint T^m(x, y) I(x, y) dx dy, \\ &= \sum_{i,j} \iint T_{ij}^m(x, y) I(x, y) dx dy, \\ &= \sum_{i,j} a_{m,q(i,j)} \iint \mathbf{1}_{E_{ij}}(x, y) I(x, y) dx dy, \\ &= \sum_{i,j} a_{m,q(i,j)} I(i, j). \end{aligned} \quad (7)$$

Equation (7) is the familiar form of compressive measurements if the pixelized image $I(i, j)$ is reordered into a vector by the mapping $q$. Indeed, in the vector form, (7) is tantamount to

$$z_m = \sum_{i,j} a_{m,q(i,j)} I(i, j) = \sum_n a_{mn} I_n, \text{ or} \quad (8)$$
$$z = A \cdot I.$$

In above, $z$ is the measurement vector, $A$ is the sensing matrix and $I$ is the vector representation of the pixelized image $I(i, j)$.

It is well known [3] that the pixelized image $I$ can be reconstructed from the measurements $z$ by, for example, solving the following minimization problem:

$$\min \|W \cdot I\|_1, \text{ subject to } A \cdot I = z, \quad (9)$$

where $W$ is some sparsifying operator such as total variation or framelets [4][5][6].

### D. Summary

To summarize, the architecture of this work can be used to make compressive measurements of the pixelized image $I$. For a given sensing matrix $A$, the entries in each row of $A$ are used to program the transmittance of the elements of the aperture assembly. With each programmed pattern for the transmittance, the sensor makes a measurement. The measurements from all rows of $A$ form a measurement vector $z$ which is given by (8). Then the measurement vector $z$ can be used to reconstruct the pixelized image $I$ from the minimization problem (9). Compressive sensing theory dictates that a good approximation of the image $I$ can be computed with far fewer measurements than the total number of aperture elements (the number of pixels of $I$). Furthermore, the more measurements are used in reconstruction, the better quality of the reconstructed image is [3].

## V. MULTI-VIEW IMAGING

Multiple sensors may be used in conjunction with one aperture assembly as shown in Figure 4. A virtual image can be defined for each sensor, say, $I^{(k)}(x, y)$ is the virtual image associated with sensor $S^{(k)}$, where the superscript $k$ is used for indexing the multiple sensors. These images are multi-view images of a same scene.

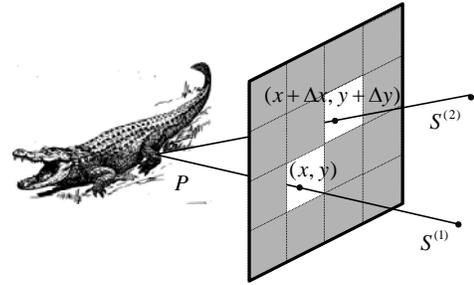

**Figure 4. Multiple sensors are used with one aperture assembly to make multi-view images**

For a given setting of transmittance $T(x, y)$, each sensor takes a measurement, and therefore, for a given sensing matrix, the sensors produce a set of measurement vectors, $z^{(k)}$, simultaneously. Each measurement vector $z^{(k)}$ can be used to reconstruct a pixelized image $I^{(k)}$ by solving problem (9) independently without taking into consideration of other measurement vectors. However, although the images $I^{(k)}$ are different, there is a high correlation between them, especially when the sensors are close to one another and when the scene is far away. The correlation between the images can be exploited to enhance the quality of the reconstructed images.

Multiple sensors with one aperture assembly may be used in the following three ways:

- In a general setting, the measurement vectors from multiple sensors represent images of different views of a scene, creating multi-view images. Thus, the architecture allows a simple device to capture multi-view images simultaneously.
- When the scene is planar, or sufficiently far away, the measurement vectors from the sensors may be considered to be independent measurements of a same image (except for small difference at the borders) and they may be concatenated as a larger set of measurements to be used to reconstruct the image. This increases number of measurements that can be taken from the same image in a given duration of time.
- When the scene is planar, or sufficiently far away, and when the sensors are properly positioned, the measurement vectors from the sensors may be considered to be the measurements made from a higher resolution pixelized image, and they may be used to reconstruct an image of the higher resolution than the number of aperture elements.

The detailed discussions will be given in the rest of this section.

*A. Image decomposition*

For simplicity, we consider two sensors, $S^{(1)}$ and $S^{(2)}$, that are placed in a same plane parallel to the plane of aperture assembly, as shown in Figure 5. The sensors define two virtual images $I^{(1)}(x,y)$ and $I^{(2)}(x,y)$. We want to explore common component between them.

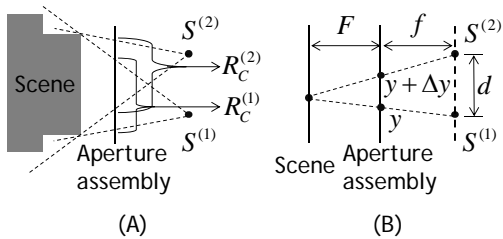

**Figure 5. Various definitions for two sensors on a plane parallel to the plane of aperture assembly. The illustration is made on a plane perpendicular to the plane of aperture assembly so that the aperture assembly is illustrated as a vertical line.**

The area of the aperture assembly can be divided into two disjoint regions, $R_C^{(1)}$ and $R_D^{(1)}$, according to $S^{(1)}$. In the simplest term, $R_C^{(1)}$ consists of the objects that can be also seen by $S^{(2)}$; that is, the objects appearing in $R_C^{(1)}$ are common in both images, $I^{(1)}(x,y)$ and $I^{(2)}(x,y)$. $R_D^{(1)}$ consists of the objects that can be only seen by $S^{(1)}$; that is, the objects appearing in $R_D^{(1)}$ can only be found in $I^{(1)}(x,y)$. The definition of the two regions can be made more precise by using the rays from the two sensors.

As shown in Figure 4, any point $(x,y)$ defines a ray that starts from the sensor $S^{(1)}$ and passes through $(x,y)$. The ray must ends at a point $P$ in the scene. Now if a ray emitted from point $P$ can reach the sensor $S^{(2)}$ through the aperture assembly without obstruction by other objects of the scene (with all aperture elements open), then $(x,y) \in R_C^{(1)}$. Otherwise, if no rays from $P$ can reach the sensor $S^{(2)}$ (with all aperture elements open), then $(x,y) \in R_D^{(1)}$. $R_C^{(2)}$ and $R_D^{(2)}$ can be similarly defined as above by reversing the role of $S^{(1)}$ and $S^{(2)}$. $R_C^{(1)}$ and $R_C^{(2)}$ are illustrated in Figure 5(A) in one dimensional view.

Incidentally, the definition of $R_C^{(1)}$ and $R_C^{(2)}$ also defines a one-to-one mapping between them. The points where the rays $\overrightarrow{PS^{(1)}}$ and $\overrightarrow{PS^{(2)}}$ intersects the aperture assembly are mapped into each other. The mapping is defined as

$$U^{12}:(x,y) \in R_C^{(1)} \mapsto (x+\Delta x, y+\Delta y) \in R_C^{(2)}$$
$$U^{21}:(x+\Delta x, y+\Delta y) \in R_C^{(2)} \mapsto (x,y) \in R_C^{(1)} \quad (10)$$

where the relationship between $(x,y)$ and $(x+\Delta x, y+\Delta y)$ is shown in Figure 4.

Now the virtual images $I^{(k)}(x,y)$ can be decomposed by using the characteristic functions of $R_C^{(k)}$ and $R_D^{(k)}$ as follows

$$I^{(k)}(x,y) = I_C^{(k)}(x,y) + I_D^{(k)}(x,y)$$
$$I_C^{(k)}(x,y) = I^{(k)}(x,y)\mathbf{1}_{R_C^{(k)}}(x,y) \quad , k=1,2. \quad (11)$$
$$I_D^{(k)}(x,y) = I^{(k)}(x,y)\mathbf{1}_{R_D^{(k)}}(x,y)$$

Furthermore, $I_C^{(1)}(x,y)$ and $I_C^{(2)}(x,y)$ are related through the following equations:

$$I_C^{(2)}(x,y) = I_C^{(1)}(U^{21}(x,y)),$$
$$I_C^{(1)}(x,y) = I_C^{(2)}(U^{12}(x,y)). \quad (12)$$

The decomposition, $I_C^{(k)}(x,y)$ and $I_C^{(k)}(x,y)$, $k=1,2$, is illustrated in Figure 6 given below.

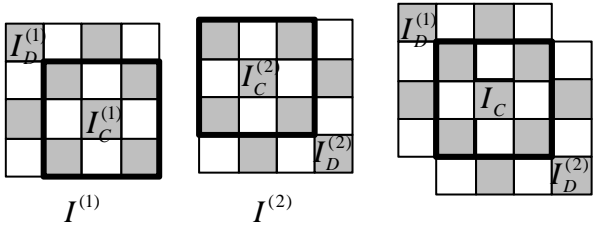

**Figure 6. Decomposition of the images from two sensors when the sensor distance is an integer multiple of the size of the aperture elements.** $I^{(k)} = I_C^{(k)} + I_D^{(k)}, k = 1, 2$. $I_C^{(1)}$ and $I_C^{(2)}$ are the common image, $I_C$, under a transform.

The significance of the decomposition (11) is that the two virtual images are decomposed into three components: one component is common to both images, and the other two components are unique to each individual image. More specifically, if we define the common component as

$$I_C(x, y) = I_C^{(1)}(x, y), \quad (13)$$

then we have

$$\begin{aligned} I^{(1)}(x, y) &= I_C(x, y) + I_D^{(1)}(x, y), \\ I^{(2)}(x, y) &= I_C(U^{21}(x, y)) + I_D^{(2)}(x, y). \end{aligned} \quad (14)$$

Since $I_C(x, y)$ is common in both images, its reconstruction may make use of the measurements from both sensors, and therefore, its quality may be enhanced as compared to only one sensor is used.

### B. Joint reconstruction

The components of the virtual images, $I_C(x, y)$, $I_D^{(1)}(x, y)$ and $I_D^{(2)}(x, y)$, can be pixelized to get three vector components $I_C$, $I_D^{(1)}$ and $I_D^{(2)}$. Referring to Figure 6, the decomposition is similar to (14) and given by

$$\begin{aligned} I^{(1)} &= I_C + I_D^{(1)}, \\ I^{(2)} &= U \cdot I_C + I_D^{(2)}. \end{aligned} \quad (15)$$

In above, $U$ is a matrix that performs shift and interpolating functions to approximate the operation of mapping $U^{21}$ defined in (10). In other words, $U \cdot I_C$ is a vector that approximates the pixelized $I_C(U^{21}(x, y))$, as given by

$$(U \cdot I_C)(q(i, j)) \approx \iint 1_{E_{ij}}(x, y) I_C(U^{21}(x, y)) dx dy. \quad (16)$$

The vector components $I_C$, $I_D^{(1)}$ and $I_D^{(2)}$ may be jointly reconstructed from the two measurement vectors, $z^{(1)}$ and $z^{(2)}$, made from the two sensors. Let $A$ be the sensing matrix with which the measurements $z_m^{(1)}$ and $z_m^{(2)}$ are made. Then the optimization problem to solve is

$$\min \|W \cdot I_C\|_1 + \frac{\sigma}{2} \sum_{k=1}^{2} \|W \cdot I_D^{(k)}\|_1, \quad \text{subject to}$$
$$A \cdot I_C + A \cdot I_D^{(1)} = z^{(1)}, \quad (17)$$
$$A \cdot U \cdot I_C + A \cdot I_D^{(2)} = z^{(2)}.$$

In (17), $\sigma > 0$ is a normalization constant to account for the areas of the four regions $R_C^{(k)}$ and $R_D^{(k)}$, $k = 1, 2$. The value of the joint reconstruction (17) lies in the fact that there are only three unknown components in (17) with two constraints (given by $z^{(1)}$ and $z^{(2)}$), as compared to four unknown components with two constraints if the images are reconstructed independently from (9). Typically, $I_C$ has much more nonzero entries than that of $I_D^{(1)}$ and $I_D^{(2)}$, hence the number of unknowns is reduced by almost a half.

In general, problem (17) is quite difficult to solve because the regions $R_C^{(k)}$ and $R_D^{(k)}$, $k = 1, 2$ are not known *a priori*, and they should be part of the solution. However, if the scene is planar and its distance is known, then it is possible to compute $R_C^{(k)}$ and $R_D^{(k)}$, $k = 1, 2$ before (17) is solved. Therefore, in such cases when $R_C^{(k)}$ and $R_D^{(k)}$ are known, problem (17) may be solved by well known established optimization process such as those in [4][5][6].

### C. Planar scene

When the scene is on a plane parallel to and with a known distance from the plane of aperture assembly, it is possible to work out explicit formulas for the mappings $U^{12}$ and $U^{21}$ of (10). As shown in Figure 5(B), let us define the distance between two sensors to be $d$, the distance between the plane of the sensors and the plane of aperture assembly to be $f$ and the distance between the scene plane and the aperture assembly to be $F$. Then the mapping $U^{12}$ is given by

$$U^{12}(x, y) = (x + \Delta x, y + \Delta y),$$
$$\sqrt{\Delta x^2 + \Delta y^2} = \frac{F}{f + F} d, \quad (18)$$
$$(\Delta x, \Delta y) \propto \overrightarrow{S^{(1)} S^{(2)}}.$$

The last line in (18) means that the two vectors have the same angle, or orientation, in their respective planes.

In general, when the scene is non-planar, equation (18) still holds, but $F$ is no long a constant. It is rather a function of position, i.e., $F = F(x, y)$, and it is also scene dependent. However, for the scene that is sufficiently far away, $F$ is large compared to $f$ so that $\frac{F}{f + F} \approx 1$, and therefore, equation (18) becomes

$$U^{12}(x, y) = (x+\Delta x, y+\Delta y),$$
$$\sqrt{\Delta x^2 + \Delta y^2} \approx d, \quad (19)$$
$$(\Delta x, \Delta y) \propto \overrightarrow{S^{(1)}S^{(2)}}.$$

According to (19), when the scene is sufficiently far away, the virtual images from the two sensors are approximately the same, except for a shift of distance $d$. Therefore, the common region $R_C^{(k)}$ covers the entire aperture assembly except for a border of width $d$. Consequently, compared to the common image $I_C$, the images $I_D^{(1)}$ and $I_D^{(2)}$ have small energy. This implies that problem (17) is mainly a problem for the single image $I_C$, while using two measurement vectors $z^{(1)}$ and $z^{(2)}$, twice as many measurements as when each of the images, $I^{(1)}$ and $I^{(2)}$, is reconstructed independently as in (9). For this reason, multiple sensors may be considered as taking independent measurements for a same image if the scene is sufficiently far away. This can be used as a mechanism to increase the number of measurements taken during a given time duration.

If the distance between two sensors, $d$, is equal to an integer multiple of the size of the aperture elements, as illustrated in Figure 6, then matrix $U$ in (17) is simply a shift matrix. In other words, the entries of $U$ are zero except for the entries on an off-diagonal, which are equal to 1.

### D. High resolution

For sufficiently far away scenes, multiple sensors may also be used as a mechanism to improve the resolution of the common image $I_C$. If the distance $d$ between two sensors is a non-integer multiple of the size of the aperture elements, then $I^{(1)}$ and $I^{(2)}$ can be considered as two down-sampled images of a higher resolution image, see Figure 7. The joint reconstruction can therefore be used to create a higher resolution image.

Specifically, equation (14) can be rewritten as
$$I^{(1)}(x, y) = I_C(x, y) \qquad\qquad + I_D^{(1)}(x, y),$$
$$I^{(2)}(x, y) = I_C(x-\Delta x, y-\Delta y) + I_D^{(2)}(x, y). \quad (20)$$

If the distance $d$ between two sensors is a non-integer multiple of the size of the aperture elements, then there is no overlapping of grid points $(x-\Delta x, y-\Delta y)$ with the grid points $(x, y)$. Therefore, equation (20) shows that images $I^{(1)}$ and $I^{(2)}$ comprise different sampling of the same image $I_C$, i.e., $I^{(1)}$ samples $I_C$ at points $(x, y)$, while $I^{(2)}$ samples $I_C$ at points $(x-\Delta x, y-\Delta y)$. Consequently, the measurement vectors $z^{(1)}$ and $z^{(2)}$ can be used to reconstruct the image $I_C$ at both grid points $(x, y)$ and $(x-\Delta x, y-\Delta y)$. This results in an image $I_C$ that has a higher resolution than given by the aperture elements. This is illustrated in Figure 7 below.

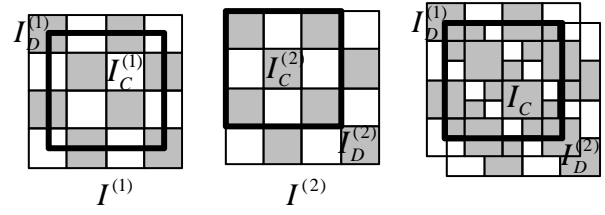

**Figure 7. Decomposition of the images from two sensors when the sensor distance is a non-integer multiple of the size of the aperture elements.** $I^{(k)} = I_C^{(k)} + I_D^{(k)}, k=1,2$. $I_C^{(1)}$ and $I_C^{(2)}$ **are the common image,** $I_C$, **under a transform.**

## VI. PRACTICAL CONSIDERATIONS

In this section, we discuss issues arising from practical implementations of the proposed architecture.

### A. Selection of aperture assembly

The architecture of this work is flexible to allow a variety of implementations for the aperture assembly.

For imaging of visible spectrum, liquid crystal sheets [13] may be used. Micromirror arrays [8] may be used for both visible spectrum imaging and infrared imaging. When a micromirror array is used, the array is not placed in the direct path between the scene and the sensor, but rather it is placed at an angle so that the rays from the scene is reflected to the sensor when the micromirrors are turned to a particular angle, see [8] for an example of arrangement. Further, when the micromirror array is used, the transmittance is binary, taking the values of 0 and 1. The metallic masks of [10][11] may be used for Terahertz imaging. For millimeter wave imaging, the mask of [12] can be used.

In all these selections, the aperture assembly is able to vary the transmittance of individual aperture element as instructed by a programmable logic.

### B. Sensor of finite size

In implementations, a sensor, such as a single photo-conductive cell, has always a finite size. We now consider the effect of a finite-size sensor. For the purpose of comparison, we use $I_f(x, y)$ to denote the virtual image from a sensor of finite size, and use $I_i(x, y)$ to denote the virtual image form an infinitesimal sensor which is located at the center of mass of the finite-size sensor. We will establish a relationship between $I_f(x, y)$ and $I_i(x, y)$.

As before, the image from a sensor of finite size is defined as the integration of all rays reaching at the sensor that pass through a point $(x, y)$ on the aperture assembly, as illustrated in Figure 8(A).

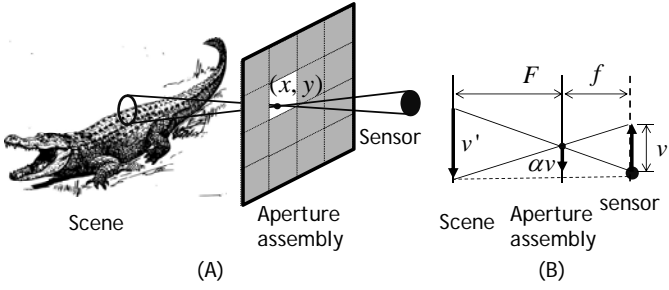

**Figure 8. Sensor of finite size. (A)** The image at point (x,y) is defined as integration of all rays within the cone passing through the point (x,y) and arriving at the sensor. **(B)** Derivation of the relationship between $I_f(x,y)$ and $I_i(x,y)$. The illustration is made on a plane perpendicular to the planes of aperture assembly and the sensor, which both appear as a line. Only upper half of the finite-size sensor (v) and the lower half of the cone of the rays are shown. The bottom of sensor (v) is where the infinitesimal sensor is located.

In this subsection, we assume the scene is on a plane parallel to the aperture assembly and has a distance of $F$ form it. We also assume the finite-size sensor has a two dimensional shape, denoted by $S$, on a plane parallel to the aperture assembly with a distance of $f$ from it, see Figure 8(B). We do not assume the sensitivity of the finite-size sensor is uniform. Let $(u,v)$ be a point on $S$, and $\rho(u,v)$ be the sensitivity of the sensor at point $(u,v)$. If the sensor has uniform sensitivity, then $\rho = \frac{1}{|S|}\mathbf{1}_S$, where $|S|$ is the area of $S$.

Referring to Figure 8(B), the upper half of the finite-size sensor is shown and labeled $v$, and the infinitesimal sensor is located at the bottom of it. The lower half of the cone of the rays from the scene that can reach at the upper half of the finite-size sensor through point $(x,y)$ is labeled by $v'$. These rays form part of the image of the infinitesimal sensor, and it is labeled $\alpha v$. The integration of the intensity of these rays is the value of $I_f(x,y)$. From the geometry shown in Figure 8(B), it can be easily verified that the factor $\alpha$ is given by

$$\alpha = \frac{F}{f+F}. \tag{21}$$

At a point $(u,v)$ on $S$, a ray on the region labeled $\alpha v$ has intensity $I_i(x-\alpha u, y-\alpha v)$, but the sensitivity of the finite-size sensor at the point is $\rho(u,v)$, and therefore, the contribution of the ray to the integral is $\rho(u,v)I_i(x-\alpha u, y-\alpha v)$. Thus, the image of finite-size sensor is given by

$$\begin{aligned}I_f(x,y) &= \iint \rho(u,v)I_i(x-\alpha u, y-\alpha v)dudv,\\ &= \frac{1}{\alpha^2}\iint \rho(\frac{u}{\alpha},\frac{v}{\alpha})I_i(x-u,y-v)dudv,\\ &= \iint \rho_\alpha(u,v)I_i(x-u,y-v)dudv,\\ &= (\rho_\alpha * I_i)(x,y),\end{aligned} \tag{22}$$

where

$$\rho_\alpha(u,v) = \frac{1}{\alpha^2}\rho(\frac{u}{\alpha},\frac{v}{\alpha}). \tag{23}$$

Equation (22) shows that the virtual image for the finite-size sensor is the convolution of the image of infinitesimal sensor with the point spread function $\rho_\alpha$, i.e., $I_f = \rho_\alpha * I_i$. In other words, the virtual image $I_f(x,y)$ of the finite-size sensor is $I_i(x,y)$ of the infinitesimal sensor, blurred by $\rho_\alpha$, whose support is smaller than the size of the finite-size sensor because the support of $\rho$ is $S$ and $\alpha < 1$.

We now consider the pixelization of $I_f(x,y)$, which is similar to (3) and given below.

$$\begin{aligned}I_f(i,j) &= \iint \mathbf{1}_{E_{ij}}(x,y)I_f(x,y)dxdy,\\ &= \iiiint \phi(x,y,u,v)dudvdxdy,\end{aligned} \tag{24}$$

$$\phi(x,y,u,v) = \mathbf{1}_{E_{ij}}(x,y)\rho_\alpha(u,v)I_i(x-u,y-v).$$

After a change of variables, equation (24) becomes

$$I_f(i,j) = \iint \kappa_{ij}(x,y)I_i(x,y)dxdy. \tag{25}$$

where

$$\begin{aligned}\kappa_{ij}(x,y) &= \iint \mathbf{1}_{E_{ij}}(x+u,y+v)\rho_\alpha(u,v)dudv,\\ &= \mathbf{1}_{E_{ij}} * \bar{\rho}_\alpha,\end{aligned} \tag{26}$$

$$\bar{\rho}_\alpha(u,v) = \rho_\alpha(-u,-v).$$

We now compare $I_i(i,j)$ and $I_f(i,j)$ when the finite-size sensor has the uniform sensitivity and when the scene is sufficiently far away from the aperture assembly. If the scene is far, then $F$ is large compared to $f$, so we can assume $\alpha = 1$. Also, $\bar{\rho}_\alpha = \bar{\rho} = \frac{1}{|S|}\mathbf{1}_S$, if the sensitivity is uniform.

Next, we rewrite the equations for $I_i(i,j)$ and $I_f(i,j)$, from equations (3) and (25), respectively as

$$\begin{aligned}I_i(i,j) &= \iint \mathbf{1}_{E_{ij}}(x,y)I_i(x,y)dxdy,\\ I_f(i,j) &= \iint \frac{1}{|S|}\left(\mathbf{1}_{E_{ij}} * \mathbf{1}_S\right)(x,y)I_i(x,y)dxdy.\end{aligned} \tag{27}$$

It is now clear from (27) that both $I_i(i,j)$ and $I_f(i,j)$ are pixelization of $I_i(x,y)$, the virtual image from the infinitesimal sensor, but the difference is that the former is integrated with $\mathbf{1}_{E_{ij}}$, and the later is integrated with $\frac{1}{|S|}\mathbf{1}_{E_{ij}} * \mathbf{1}_S$. In other words, while $I_i(i,j)$ is obtained from a pixelization of disjoint regions defined by the aperture elements, $E_{ij}$, $I_f(i,j)$ is the result of pixelization by overlapped regions, resulting in blurring. The overlapped regions are determined by $S$, the shape of the sensor. The blurring is negligible if the area of $S$ is much smaller than that of $E_{ij}$, that is, if the sensor is much smaller than the aperture element.

The equation for $I_f(i,j)$ in (27) is useful in reconstruction of an image when using measurements from a finite-size sensor. In the reconstruction, we should try to reconstruct, not $I_f(i,j)$, but some discretized version of $I_i(x,y)$ by using the constraints consistent to how the measurements are actually obtained. For example, let $z$ be the measurement vector obtained with a finite-size sensor by using the sensing matrix $A$. Then we reconstruct an image by solving the following problem of finding the image $I_i(x,y)$ of the infinitesimal sensor:

minimize a cost function of $I_i(x,y)$, subject to

$$z_m = \frac{1}{|S|}\sum_{i,j} a_{m,q(ij)} \iint \left(\mathbf{1}_{E_{ij}} * \mathbf{1}_S\right)(x,y) I_i(x,y) dx dy, \quad (28)$$

as opposed to solving the conventional problem of finding the image $I_f(i,j)$ of the finite-size sensor :

minimize a cost function of $I_f(i,j)$, subject to

$$z_m = \sum_{i,j} a_{m,q(ij)} I_f(i,j). \quad (29)$$

The solution to the conventional problem (29) would result in a blurring due to the finite size of the sensor. However, by solving the minimization problem (28) in reconstruction, the effect of the finite-size sensor is accounted for, and the blurring is removed.

It is worthwhile to point that the blurring given in (25) due to the finite size of the sensor is different from the blurring due to objects being out of focus of a lens. The blurring in (25) does not exist if an infinitesimal sensor can be built, but it is still possible for an object to be out of focus even if a theoretically perfect lens is built. The blur in (25) is caused by the inability to make an infinitesimal sensor, which is analogous to the fact that an image created by a realistic lens can never be perfectly in focus because it is impossible to built a theoretically perfect lens.

*C. Super-resolution*

When the sensor has an infinitesimal size, the resolution of the reconstructed image is the same as the resolution of the aperture assembly, as shown in (3) and (9). However, for a finite-size sensor, an image may be reconstructed with a different resolution than that of the aperture assembly. In particular, with a finite-size sensor, it is theoretically possible to reconstruct an image of a resolution much higher than the resolution of the aperture assembly.

Equation (28) provides a method to reconstruct a virtual image $I_i(x,y)$ which can be considered to have infinite resolution, because it is a function of continuous variables $(x,y)$. However, it is not expected that the constraints in (28) are able to determine a unique $I_i(x,y)$ for continuous variables $(x,y)$ in general, because there is only a finite number of constraints. On the other hand, if there is some prior knowledge of $I_i(x,y)$, super-resolution reconstruction is possible. For example, if it is known that the image is created by a point lighting source, and if the sensor has the same size and shape as the aperture elements, it is theoretically possible to find the exact location of the point source by solving problem (28).

Equation (28) is also flexible in allowing pixelization of different granularity in different regions of an image, for example, multi-resolution. The pixelization can be done by 1) dividing the image into small regions which are called pixels (the regions may have different shapes or sizes), and 2) assuming that $I_i(x,y)$ is a constant in each of the pixel regions (the constant is the pixel value at the pixel). Then the integrals in (28) may be calculated to yield a set of constraints on the pixel values.

*D. Diffraction of aperture*

In implementations, when the aperture elements are small, the effect of diffraction must be considered. For this purpose, we consider an image from an infinitesimal sensor, of a monochromatic wave with the wave number $k$. Let $I_d(x,y)$ be the virtual image with diffraction effect when the aperture assembly has the transmittance $T(x,y)$. As before, let $I_i(x,y)$ be the virtual image without the effect of diffraction. Then $I_d(x,y)$ can be written in terms of $I_i(x,y)$ by Fraunhofer diffraction equation [14] as

$$I_d(x,y) = \iint e^{-\sqrt{-1}k(lu+hv)} T(u,v) I_i(u,v) du dv. \quad (30)$$

In (30), $l, h$ are the direction cosines of the point $(-x,-y)$ with respect to the origin which is located at the infinitesimal sensor. Equation (30) shows that the effect of diffraction causes a blur in the image, much as the effect of finite-size sensor does in (22), but of course, with a different point spread function. Furthermore, since $T(x,y)$ is involved in (30), the blur caused by the diffraction actually depends on the pattern of the aperture assembly.

Now, let the transmittance of aperture element $E_{ij}$ be $T_{ij}$ and further assume that for any two aperture elements, $E_{ij}$ and $E_{st}$, the following integral is a constant over $E_{st}$:

$$T_{st} \iint_{E_{ij}} e^{-\sqrt{-1}k(lu+hv)} dxdy = const, \ (u,v) \in E_{st}, \quad (31)$$
$$\triangleq b_{q(i,j),q(s,t)}.$$

Then we can derive a simple relationship between the pixelized images with and without diffraction. Indeed, the pixelized image with diffraction is given by

$$\begin{aligned}
I_d(i,j) &= \iint_{E_{ij}} I_d(x,y)dxdy, \\
&= \iint_{E_{ij}} \iint e^{-\sqrt{-1}k(lu+hv)} T(u,v) I_i(u,v) dudvdxdy, \\
&= \sum_{s,t} T_{st} \iint_{E_{ij}} \iint_{E_{st}} e^{-\sqrt{-1}k(lu+hv)} I_i(u,v) dudvdxdy, \quad (32) \\
&= \sum_{s,t} b_{q(i,j),q(s,t)} \iint_{E_{st}} I_i(u,v) dudv, \\
&= \sum_{s,t} b_{q(i,j),q(s,t)} I_i(s,t).
\end{aligned}$$

In the vector form, the pixelized image can be written as
$$I_d = B \cdot I_i, \quad (33)$$

where $B$ is a square matrix with entries $b_{q(i,j),q(s,t)}$ defined in (31). Equation (33) shows that the effect of the diffraction is simply a blurring with the kernel matrix $B$ whose entries are given in (31). If a measurement $z_m$ is made by using a row vector $a^{(m)}$ of a sensing matrix $A$, then in the reconstruction, the measurement needs to be considered as if made by the modified row vector $a^{(m)} B^{(m)}$ in order to account for the diffraction effect. Note that the superscript $m$ is used in $a^{(m)} B^{(m)}$ because matrix $B$ in (33) actually depends on the pattern of the aperture assembly when measurement $z_m$ is made.

## VII. PROTOTYPE

In this section, we describe the prototype and present examples from the prototype device.

The imaging device consists of a transparent monochrome liquid crystal display (LCD) screen and a photovoltaic sensor enclosed in a light tight box, shown in Figure 9. The LCD screen functions as the aperture assembly while the photovoltaic sensor measures the light intensity. The photovoltaic sensor is a tricolor sensor, which outputs the intensity of red, green and blue lights. A computer is used to generate the patterns for aperture elements on LCD screen according to each row of the measurement matrix. The light measurements are read from the sensor and recorded for further processing. The computer is also responsible for synchronization between the creation of patterns on the LCD and the timing of measurement capture, see Figure 10.

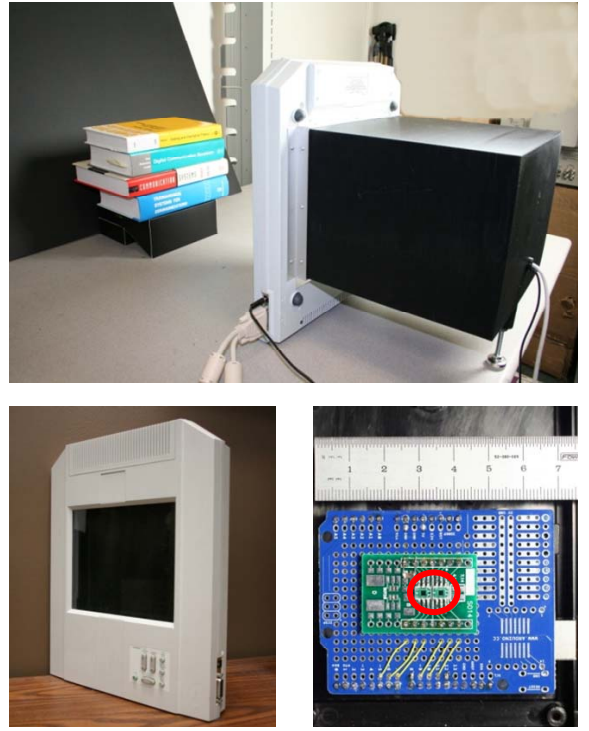

**Figure 9. Prototype device. The top photo shows the laboratory set up to acquire the image of the books. The bottom left is the LCD screen used as the aperture assembly, and the bottom right is a photo of the RGB sensor board. Two sensors, indicated by the red circle, are mounted on the board.**

### A. Image acquisition

The LCD panel is configured to display a maximum resolution of 302 x 217 = 65534 black or white squares. Since the LCD is transparent and monochrome, a black square means the element is opaque, and a white square means the element is transparent. Therefore, each square represents an aperture element with transmittance of a 0 (black) or 1 (white).

For capturing compressive measurements, we use a sensing matrix which is constructed from rows of a Hadamard matrix of order $N$=65536. Each row of the Hadamard matrix is permuted according to a predetermined random permutation. The first 65534 elements of a row are then simply mapped to the 65534 aperture elements of the LCD in a scan order from the top to bottom and then from left to right. An '1' in the Hadamard matrix turns an aperture element transparent and a '-1' turns it opaque. The measurements values for red, green and blue are taken by a sensor at the back of the enclosure box and recorded by the control computer, as illustrated in Figure 10.

In experiments reported in this paper, only one sensor is used to take the measurements. Results for multi-view imaging with two sensors will be reported in a future paper.

A total number of 65534, which corresponds to the total number of pixels of the image, different measurements can be made with the prototype. In our experiments, we only make a fractional of the total possible measurements. We express the number of measurements taken and used in reconstruction as a percentage of the total number of pixels. For example, 50% of measurements means 32767 measurements are taken and used in reconstruction, which is half of the total number of pixels, 65534. Similarly, 25% means 16384 measurements are taken and used in reconstruction.

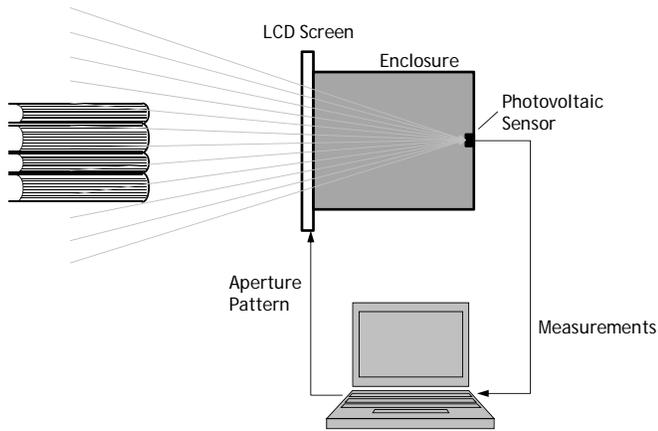

**Figure 10. Schematic illustration of the lensless compressive sensing image prototype.**

*B. Image Reconstruction*

We used various still life subjects in the laboratory to demonstrate the concept of the imaging device. We rely on the standard reconstruction method commonly known as L1 minimization of total variation by solving Eq. (9).

The number of measurements needed for reconstruction of an image depends on many factors such as the complexity (features) of the image and quality of the reconstructed image. Figure 11 shows reconstructed images of a soccer ball with 12.5% and 50% measurements.

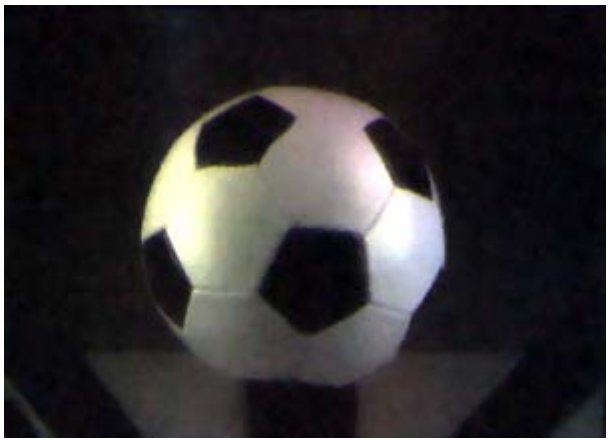

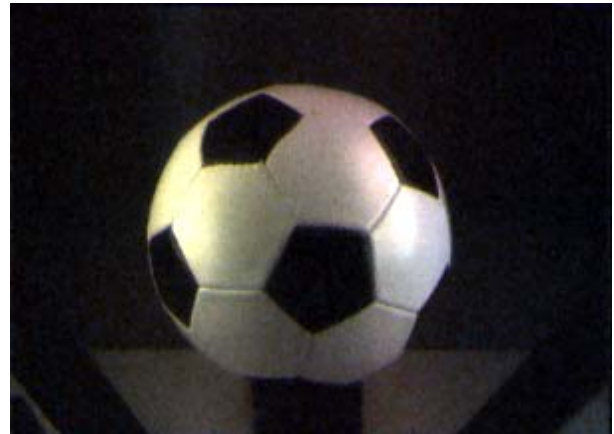

**Figure 11. Reconstructed images of "Soccer". Top: 12.5% measurements. Bottom: 50% measurements.**

Figure 12 shows reconstructed images with relatively more features. The reconstruction of the images used 25% and 50% of total measurements, respectively. Figure 13 shows reconstructed images of a cat sleeping in a basket with 25% and 50% of total measurements.

We note that the color images are reconstructed by using directly the measurements of the three color components from the sensor. No calibrations were made to balance the color components.

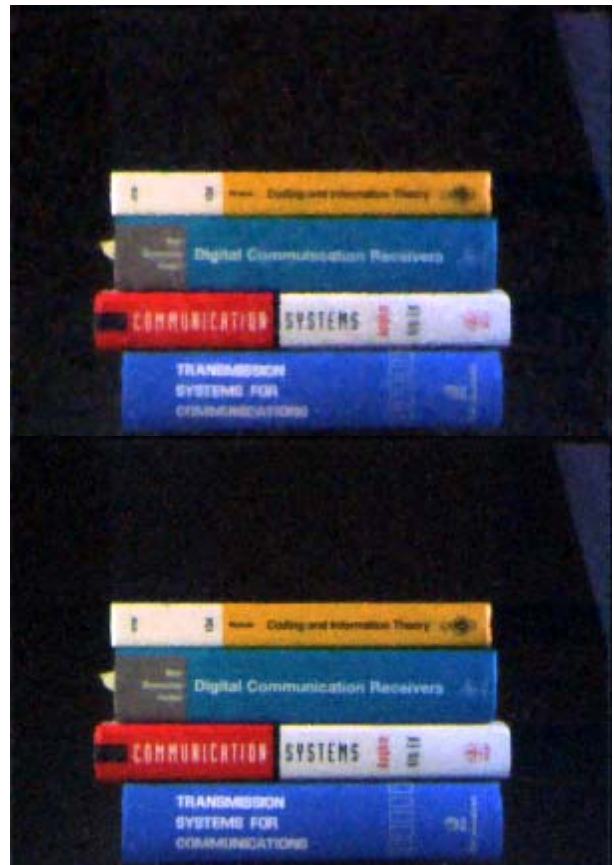

**Figure 12. Reconstructed images of "Books". Top 25% measurements. Bottom 50% measurements**

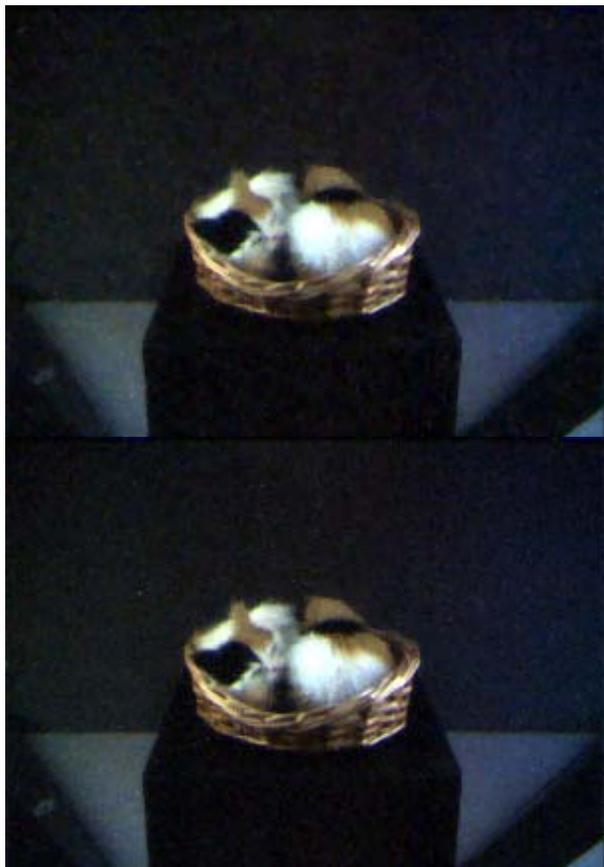

**Figure 13. Reconstructed images of "Sleeping cat". Top: 25% measurements. Bottom: 50% measurements.**

## VIII. Conclusion

An architecture for lensless compressive sensing imaging is proposed. The architecture allows flexible implementations to build simple, reliable imaging devices with reduced size, cost and complexity. Furthermore, the images from the architecture do not suffer from such artifacts as blurring due to defocus of the lens. Devices based on this architecture may be used in surveillance applications for detecting anomalies or extracting features such as speed of moving objects.

Discussion and analysis were presented on how to handle multi-view images efficiently, how to deal with the effects of finite-sized sensor and diffraction, and how to reconstruct images with higher resolution.

A prototype device was built using low cost, commercially available components to demonstrate that the proposed architecture is indeed feasible and practical.

## Acknowledgment

We thank Hock Ng and Bob Farah of Alcatel-Lucent for helping with the prototype enclosure. We also thank Raziel Haimi-Cohen, Songqing Zhao and Larry O'Gorman of Alcatel-Lucent for their interests and fruitful discussions leading to improvement of this paper.